\documentclass[journal,onecolumn,draftnocls,12pt,twoside]{IEEEtran}

\usepackage{times}
\usepackage{epsfig}
\usepackage{graphicx}
\usepackage{amsmath}
\usepackage{amssymb}
\usepackage{enumitem} 
\usepackage{graphicx}
\usepackage{subcaption}
\usepackage{url}
\usepackage{cite}

\begin{document}
\title{GSANet: Semantic Segmentation with \\ Global and Selective Attention}

\author{Qingfeng Liu, Mostafa El-Khamy, Dongwoon Bai, Jungwon Lee
\thanks{The authors are with Samsung SOC R\&D Lab, San Diego, CA 92121, USA}
\thanks{Email: m\_elkhamy@ieee.org}
}

%\ninept
%
\maketitle
\begin{abstract}
This paper proposes a novel deep learning architecture for semantic segmentation. 
The proposed Global and Selective Attention Network (GSANet) features Atrous Spatial Pyramid Pooling (ASPP) with a novel sparsemax global attention and a novel selective attention that deploys a condensation and diffusion mechanism to aggregate the multi-scale contextual information from the extracted deep features. A selective attention decoder is also proposed to process the GSA-ASPP outputs for optimizing the softmax volume.
We are the first to benchmark the performance of semantic segmentation networks with the low-complexity feature extraction network (FXN) MobileNetEdge, that is optimized for low latency on edge devices. We show that GSANet can result in more accurate segmentation with MobileNetEdge, as well as with strong FXNs, such as Xception. GSANet improves the state-of-art semantic segmentation accuracy on both the ADE20k and the Cityscapes datasets. 
%In particular on ADE20k dataset, single model can achieve 47.27\% mIoU which is 1.6\% improvement than the state-of-the-art.

\end{abstract}
%
%\begin{keywords}
%GSANet, Selective Attention ASPP, Global Attention Feature, sparsemax
%\end{keywords}
%
\section{Introduction}
\label{sec:intro}

Semantic segmentation is an important computer vision task that has many industrial applications, like autonomous driving, medical image analysis and mobile phone cameras. 
There has been significant progress by recent research works that proposed deep neural networks for semantic segmentation. The accuracy of pixel-level semantic segmentation improves with the knowledge of multiscale contextual information. PSPNet ~\cite{pspnet} performs spatial pyramid pooling (SPP) at several grid scales. Atrous spatial pyramid pooling (ASPP) has been inspired by SPP and utilized in DeepLabV3+~\cite{deeplabv3+}. ASPP collects information from the extracted features at different scales and receptive fields using dilated convolutions with different dilation rates. Another approach, called self-attention~\cite{vaswani2017attention} gathered momentum, as it has been shown it can capture the important information without being constrained to a local regular grid. 
Self-attention ~\cite{fu2019dual, ocnet} and its low cost variants ~\cite{huang2019ccnet, zhu2019asymmetric} have been adopted for semantic segmentation.

We address in this paper two important issues that are crucial to more accurate semantic segmentation, and that have previously received little or no attention.
The first issue is how to fuse the captured multiscale contextual information. DeepLabV3+~\cite{deeplabv3+} combines the contextual information using a simple 
$1\times1$ convolutional filter that cannot model the relative importance of the different scales at the location of interest. The second issue is how to obtain relevant global contextual information. For example, the ASPP deployed in DeepLabV3 deploys global average pooling (GAP) to capture the mean feature as the global contextual information of the input feature, which will be the same at all spatial locations. 
%However, each pixel might rely on different global contextual information rather than an unified single global contextual information for all the pixels.

This paper proposes a novel architecture called the Global and Selective Attention Network (GSANet) to address these issues. GSANet constitutes of a novel Selective Attention (SA) module for the fusion of the contextual information from the multiple scales, while giving the appropriate relative importance to the different scales at the different spatial locations. The proposed SA-ASPP module aggregates multiscale contextual information from the Feature eXtraction Network (FXN) using atrous convolutions with different dilation rates. The proposed selective attention combines these features by calculating the relative importance of the multiscale features using a condensation and diffusion mechanism. GSANet also features a Global Attention Feature (GAF) module (instead of global average pooling) to provide a more relevant global feature with different global contextual information at the different spatial locations, and reflect the different perceptions of the global information when observed from different locations.  
Another contribution is Sparsemax-GAF that deploys sparsemax normalization (instead of the popular softmax normalization) when calculating the GAF so as to suppress the noisy long range contextual information and boost the more prominent global information. The proposed GSA-ASPP calculates the sparsemax-GAF as the global features, and combines this global feature with the multiscale features extracted at different dilation rates using the SA.
Please note that the selective attention module and the GAF module are not limited to the context of ASPP. 
They can serve to enhance any network modules that requires calculation of global contextual information and the fusion of multiscale information. 
Moreover, we propose the selective attention decoder (SA-Dec), that uses selective attention to combine the GSA-ASPP output together with some low-level features extracted by the FXN for better decoding.

Other contributions of this paper is the investigations of segmentation networks for best performance, as well as for efficiency. With the Xception network as the FXN, we show that the GSANet provides the state-of-the-art (SOTA) accuracy on public datasets ADE20k and Cityscapes.   We are also first to benchmark the performance of the light network MobileNetEdgeTPU~ \cite{mobilenetEdgeTPU} (MNEdge), that has been optimized for fast performance on the edge devices, in the semantic segmentation task. We show that with both the MNEdge FXN and the Xception FXN,  GSANet provides more accurate semantic segmentation predictions than the popular DeepLabV3+ framework.

\begin{figure}[tb]
  \centering
  \includegraphics[width=\textwidth]{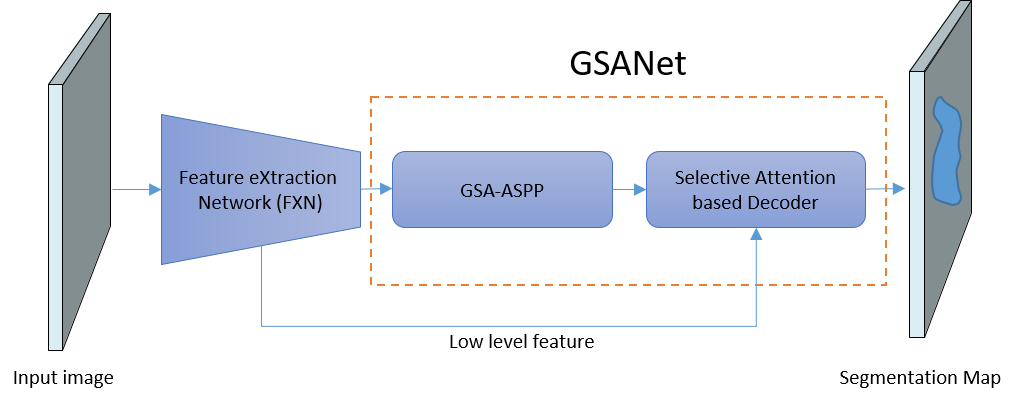}
  \caption{The system architecture of the proposed GSANet.}
  \label{fig:system_architecture}
\end{figure}

The rest of this paper is organized as follows. Section~\ref{sec:GSA} describes our proposed GSANet and its novel components: 
the SA module in Section~\ref{sec:s-aspp}, the sparsemax-GAF module in Section~\ref{sec:sparsemax-GAF}, and the SA-Dec in Section~\ref{sec:decoder}. Our experimental setup, implementation details, results, and ablation studies are presented in Section~\ref{sec:experiments}. Conclusions and discussions are made in Section~\ref{sec:conclusion}.

%%The whole system architecture of the proposed GSANet for semantic segmentation is illustrated in Figure \ref{fig:system_architecture}.
%The contributions of the paper are summarized as follows.
%\begin{itemize}
%\item This paper proposed a selective attention module and applied it to ASPP for better fusion of multiple scale information.
%\item This paper proposed a novel Global Attention Feature module using sparsemax based spatial self attention to learn a better global contextual information.
%\item This paper proposed a Selective Attention Decoder that can improve the performance.
%\item The proposed method achieved state-of-the-art performance on public benchmark and we are the first to benchmark the performance of semantic segmentation based on the lightweight feature extraction network, namely MobileNetEdgeTPU \cite{mobilenetEdgeTPU}, that are suitable for deployment on edge devices.
%\end{itemize}

\section{Global and Selective Attention Network } \label{sec:GSA}
\begin{figure}[tb]
  \centering
  \includegraphics[width=\textwidth]{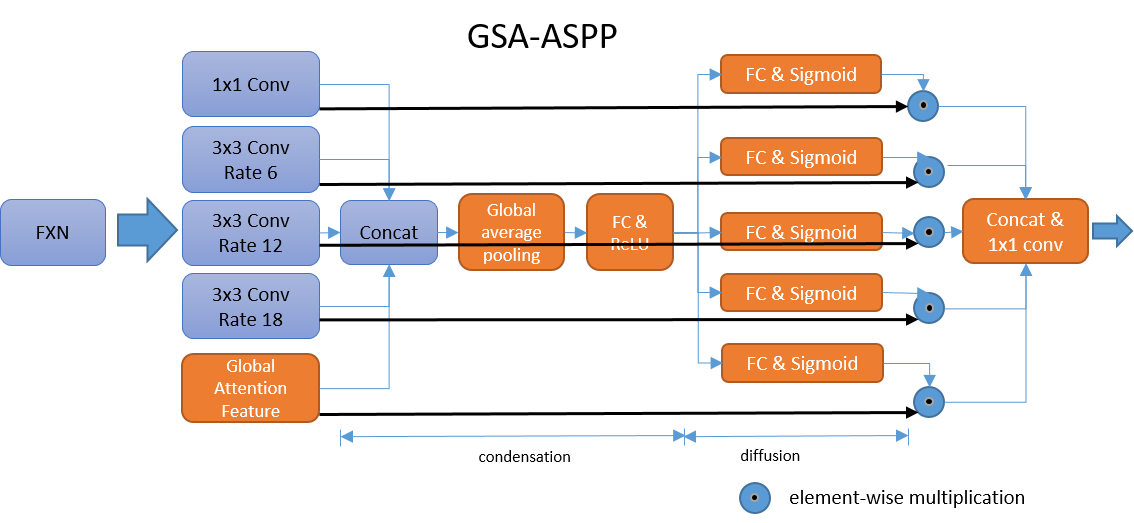}
  \caption{The architecture of GSA-ASPP. It contains two modules, namely the SA-ASPP and sparsemax-GAF. FXN is the Feature eXtraction Network, such as MobileNetEdgeTPU etc.}
  \label{fig:gsa_aspp}
\end{figure}

The proposed GASNet architecture is demonstrated in Fig.~\ref{fig:system_architecture}.
First, deep features are extracted using the FXN. Any deep neural network which showed good accuracy on the classification task, can be used as the FXN. In this paper, we show results when the FXN is either the Xception or the MNEdge network. The extracted deep features are processed by the GSA-ASPP module, shown in Fig.~\ref{fig:gsa_aspp}, which is composed of two sub-modules, namely the SA-ASPP and the sparsemax-GAF module, illustrated in Fig.~\ref{fig:gsa_aspp} and Fig.~\ref{fig:sparsemax-GAF}, respectively.

\subsection{Selective Attention ASPP (SA-ASPP)} \label{sec:s-aspp}
The ASPP has multiple atrous filters with different dilation rates, as well as  global filter, to aggregate dense information from the FXN features at different scales and receptive fields. This is crucial for pixel-level semantic segmentation, where it is need to understand the local and global context surrounding the pixel of interest, in order to classify it.
The SA-ASPP is proposed to  also use the context information, to give different spatial weights to these aggregated multiscale features, before combining them.  
First, in the condensation stage, all the $W\times H\times C$ dense features obtained from the atrous and global filters are concatenated along the feature dimension to form a $W\times H\times nC$ feature (for the case of $n-1$ atrous filters and 1 global filter) , and then condensed from all scales across the spatial and channel dimensions with global average pooling (GAP) followed by two fully connected layers with a ReLu nonlinearity to form the $1\times 1\times \gamma$ condensate feature.  
In the diffusion stage, $n$ channel attentions are obtained from the condensate feature using $n$ separate fully connected layers, to diffuse the attention back to each branch. 
Due to the condensation and diffusion mechanism, it is easy to see that the channel attention for each branch does not only use the global contextual information from its own branch, but also borrows information from multiple branches.
In Section \ref{sec:abalation}, we have show that both the condensation and diffusion stages are necessary to obtain better performance. For combining the different multiscale features, the element-wise dot product is applied per channel with the attention values and all the weighted branches are concatenated again to produce the new concatenated feature for further processing.

%Specifically, the following steps are conducted for SA-ASPP.

%
%\begin{itemize}
%\item (1)	First, the dilated convolution branches of different dilated rates and the global average pooling are concatenated to form a concatenated feature.
%\item (2)	Second, the global average pooling is applied on the concatenated feature to capture the global contextual information for each channel.
%\item (3)	Then a fully connected layer followed by ReLU activation function is used to reduce the dimension and introduce non-linearity.
%\item (4)	Then, several fully connected layers are applied on the previous dimension-reduced fully connected layer to go back to original dimension which is the same as the channel size. They are all normalized by sigmoid function, the obtained values will serve as the attentions for all the channels in each branch. 
%\item (5)	Finally, the element-wise dot product is applied per channel with the attention values and all the weighted branches are concatenated again to produce the new concatenated feature for further processing.
%\end{itemize}

\subsection{Global Attention Feature (GAF)} \label{sec:sparsemax-GAF}

The perception and understanding of the global context is crucial for accurate dense prediction. 
The ASPP~\cite{deeplabv3+} utilizes global average pooling (GAP) to extract the global contextual information that will be uniformly used at different spatial locations. One drawback of GAP is that it treats all pixels similarly. However, pixels at different spatial locations and belonging to different categories should perceive different global contextual information. Our proposed GAF is a global feature and addresses this issue by using self-attention. The global context for each pixel is derived from a linear combination of the features at all the other pixels using a distance metric determined by the degree of similarity between the pixels. If this feature is normalized by the ubiquitous softmax, we will refer to this as softmax-GAF. To give more weights to important information and discard the noisy information, we utilize the sparsemax projection~\cite{martins2016softmax} instead of softmax normalization to form the sparsemax-GAF. Sparsemax projection  performs Euclidean projection of the input attention vector $\textbf{z}=[z_1, ..., z_K]$ onto a probability simplex, as described mathematically below, with the notation that $z_{(1)} \geq z_{(2)}, ..., \geq z_{(K)}$ after sorting $\textbf{z}$ and 
\begin{equation}
	\text{sparsemax}(z_i) = \max(0, z_i - \tau(\textbf{z})),
	\label{equ:sparsemax}
\end{equation} 
where the threshold $\tau(\textbf{z})$ is found by $\tau(\textbf{z}) = \frac{\left(\sum_{j \leq f(\textbf{z}) }z_{(j)} \right)-1}{f(\textbf{z})}$,
$$f(\textbf{z}) = \max_{k \in \{1,2,..,K\}} \left( 1 + kz_{(k)} > \sum_{j \leq k} z_{(j)} \right).$$
Due to the projection and thresholding, Sparsemax produces sparse probabilities that lead to a selective and more compact attention focus. The sparsemax-GAF  strengthens the similarities and amplifies the attentions for similar pixels, while forcing zero attentions on dissimilar pixels and discarding noisy features.

\begin{figure}[tb]
  \centering
  \includegraphics[width=\textwidth]{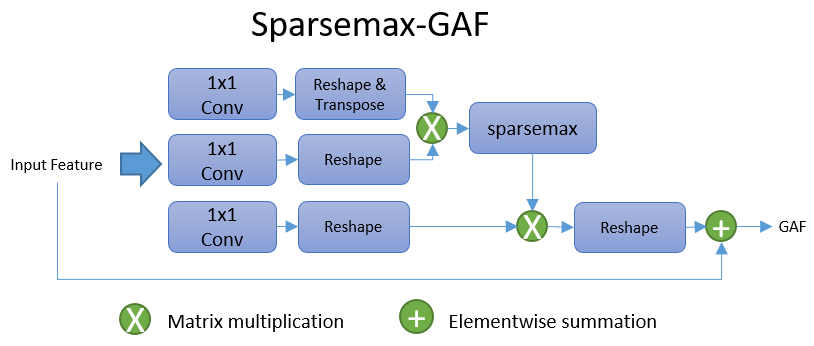}
  \caption{The architecture of the network proposed to generate the Sparsemax Global Attention Feature (sparsemax-GAF).}
  \label{fig:sparsemax-GAF}
\end{figure}

The overall pipeline of the proposed GAF module is demonstrated in Fig.~\ref{fig:sparsemax-GAF}.
First, the pairwise correlations are calculated for each pixel pair in the input feature map. Then, the raw attention map is calculated, where each value in the map corresponds to the correlation between two pixels. For sparsemax-GAF, the sparsemax projection is applied on each row of the map to obtain the normalized attention map. Third, the normalized attention map is projected onto the original input feature map using matrix multiplication to obtain the attended feature. The GAF is calculated as  the sum of the shortcut input feature and the attended feature.

\subsection{Selective Attention Decoder (SA-Dec) } \label{sec:decoder}
\begin{figure}[tb]
  \centering
  \includegraphics[width=\textwidth]{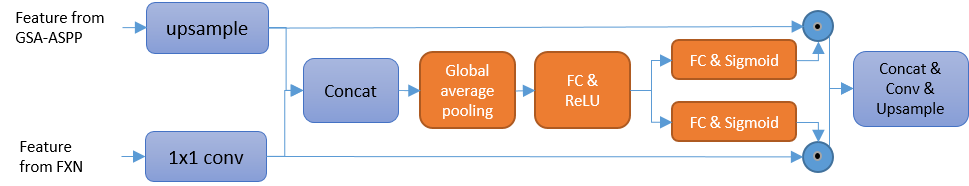}
  \caption{The network architecture of the proposed Selective Attention Decoder (SA-Dec).}
  \label{fig:s_decoder}
\end{figure}

The proposed SA-Dec builds on top of the DeepLabV3+ decoder. The SA module is inserted to selectively combine, with attention, the different features that are input to the decoder.   As shown in Fig.~\ref{fig:s_decoder}, the SA-Dec calculates the selective attention for two features, the low-level feature from the FXN and the multiscale aggregated feature that is output from the GSA-ASPP. The SA module follows the same condensation and diffusion method describes in Sec.~\ref{sec:s-aspp} to modify the two features, before being processed by the DeepLabV3+ decoder operations constituting concatenation, filtering with convolutional layers, and upsampling.  

\section{Experiments}\label{sec:experiments}

In this section, extensive experiments are conducted to shown the effectiveness of the proposed GSANet semantic segmentation architecture with different feature extraction networks.

\subsection{Datasets}
\label{ssec:datasets}

We benchmarked the performance of GSANet on two widely used public datasets, namely the Cityscapes and the ADE20k dataset.

\textbf{Cityscapes} ~\cite{cityscapes} is specifically created for scene parsing.
There are 5k high quality finely annotated images and 20k coarsely annotated images, which are taken on the street road with vehicles.
The size of all the images in the dataset is $2048\times 1024$. 
The finely annotated images are divided into 2975, 500, and 1525 splits for training, validation, and testing, respectively. 
The dataset contains 30 classes annotations in total, while only 19 classes are used for evaluation. 

\textbf{ADE20K} ~\cite{ade20k} dataset is a large-scale dataset used in ImageNet Scene Parsing Challenge 2016. 
For the challenging version, there are 150 classes and the dataset is divided into 20k, 2k, and 3k images for training, validation, and testing, respectively.  
Most of the images in the dataset are taken from real life scenes and full of diversity, including objects and scenes of various scales, shapes, and colors etc..
Different from Cityscapes, both scenes and stuff are annotated in this dataset, posing more challenges to participated methods. 
	
\subsection{Implementation Details}
\label{sec:implementation}

We demonstrate the effectiveness of the GSANet semantic segmentation architecture 
 with a low-complexity FXN, the MobileNetEdgeTPU (MNEdge), and show it can achieve competitive performance.
We also demonstrate that GSANet beats current SOTA performance with the stronger Xception FXN. 

The FXNs namely MNEdge and Xception, are pretrained on the ImageNet ~\cite{Imagenet}.
We use Stochastic Gradient Descent (SGD) to optimize our network, in which we set the initial learning rate to 0.01 for Cityscapes and 0.007 for ADE20K. 
During training, the learning rate is decayed according to the ``poly" leaning rate policy, where the learning rate is multiplied by $1 - (\frac{\mathrm{iter}}{\mathrm{max\_iter}})^{\mathrm{power}}$ with $\mathrm{power}=0.9$. 
For Cityscapes, we randomly crop out half-resolution patches $512 \times 1024$ from the original images as the inputs. 
While for ADE20K, we set the crop size to $480 \times 640$. 
For all datasets, we apply random scaling in the range of [0.5, 2.0], random horizontal flip as additional data augmentation methods.
For the ablation studies below, half resolution image of the Cityscapes dataset, namely $512 \times 1024$ are taken as the input for models using the MNEdge FXN, and the FLOPs are calculated correspondingly. At the inference stage, we use single scale inference for models using the MNEdge FXN.
For Xception based model, we adopted the left-right flipping and multiscale [0.75, 1.0, 1.25, 1.5, 1.75, 2.0] strategies for inference, following the methodology adopted by the state-of-the-art methods (\emph{cf.} DeepLabV3+) for fair comparisons.

% \subsubsection{Abalation Study on GSANet}
% \label{ssec:aba_s-aspp}

\begin{table}[tb]
  \centering
  \small
  \begin{tabular}{p{2.0cm}|c|c|c}
    \hline
    Method & mIoU (\%) & \# Params (M) & FLOPs (B) \\ \hline
    MNEdge + ASPP  & 70.45 & 3.15 & 26.77 \\ \hline
    MNEdge + SA-ASPP (condensation only) & 71.08 & 3.20 & 26.74  \\ \hline
    MNEdge + SA-ASPP (diffusion only) & 71.10 & 3.18 & 26.76  \\ \hline
    MNEdge + SA-ASPP & 71.21 & 3.20 & 26.77  \\ \hline
    MNEdge + softmax-GAF-ASPP & 70.48 & 3.20 & 27.14  \\ \hline
    MNEdge + sparsemax-GAF-ASPP & 71.19 & 3.20 & 27.18  \\ \hline
    MNEdge + GSA-ASPP & 72.10 & 3.20 & 27.18  \\ \hline
    MNEdge + GSANet & \textbf{75.07} & 3.43 & 37.64 \\ \hline
  \end{tabular}
  \caption{Ablation study of GSANet on Cityscapes validation dataset using MobileNetEdgeTPU (MNEdge). The inference input resolution is half the resolution of the Cityscapes dataset, namely $512 \times 1024$.}
  \label{table:gsanet_abalation}
\end{table}
%\subsection{Comparison with the State-of-the-art}
%\label{ssec:comparison}

%In order to show the effectiveness of the proposed modules, we also compare our method with the state-of-the-art methods using a strong FXN, namely Xception-65.

\begin{table}[tb]
  \centering
  \begin{tabular}{l|c}
    \hline
    Method & mIoU (\%) \\ \hline
    EncNet ~\cite{encnet}  & 44.65  \\ \hline
    CCNet ~\cite{huang2019ccnet} & 45.22 \\ \hline
    APNB ~\cite{zhu2019asymmetric} & 45.24 \\ \hline
    OCNet ~\cite{ocnet} & 45.45 \\ \hline
    Xception65-DeepLab V3+ ~\cite{deeplabv3+}  & 45.65  \\ \hline
    Xception65-GSANet (ours) & \textbf{47.20}  \\ 
    \hline
  \end{tabular}
  \caption{Comparison to state-of-the-art on the validation set of ADE20K.}
  \label{table:ade20k_comparison}
\end{table}

\begin{table}[tb]
  \centering
  \begin{tabular}{l|c}
    \hline
    Method & mIoU (\%) \\ \hline
    CCNet ~\cite{huang2019ccnet} & 81.3 \\ \hline
    DANet ~\cite{fu2019dual} & 81.50 \\ \hline
    ACFNet ~\cite{zhang2019acfnet} & 81.46 \\ \hline
    Xception65-DeepLabV3+ ~\cite{deeplabv3+}  & 80.42  \\ \hline
    Xception65-GSANet (ours) & \textbf{82.04}  \\ 
    \hline
  \end{tabular}
  \caption{Comparison to state-of-the-art on the validation set of Cityscapes.}
  \label{table:cityscapes_comparison}
\end{table}

\subsection{Abalation Studies and Comparisons with SOTA Methods}
\label{sec:abalation}
Table \ref{table:gsanet_abalation} shows the effectiveness of GSANet with a low-complexity FXN, MNEdge. 
The output stride, namely the ratio of the input image size over the size of the FXN output feature,  in these experiments is 16.
SA-ASPP (condensation only) means we don't use the diffusion part using multiple FC layers, but use a single FC layer instead to obtain channel attentions.
SA-ASPP (diffusion only) means no concatenation and global average pooling are used, and each branch has its own channel attention.
Compared to the DeepLabV3+ ASPP, the results show SA-ASPP provides $1.5\%$ point gain in the mean intersection over union (mIoU) accuracy.
 Table \ref{table:gsanet_abalation} also shows that sparsemax-GAF with the GSA-ASPP has $0.7\%$ point improvement in mIoU  over the GAP in SA-ASPP.
%Please note that the softmax-GAF and sparsemax-GAF refer to the softmax and sparsemax normalization of the global attention map, respectively.
We can also observe that the sparsemax-GAF gives better mIoU ($0.7\%$ point gain) than softmax-GAF.

Table \ref{table:ade20k_comparison} and Table \ref{table:cityscapes_comparison} benchmarks GSANet with the strong FXN, Xception-65, on ADE20k and Cityscapes, respectively. Xception65-GSANet provides around $1.6\%$ gain over Xception65-DeepLab V3+ on both the ADE20k and the Cityscapes datasets.

\section{Conclusions and Discussions}
\label{sec:conclusion}

This paper proposes a novel semantic segmentation architecture, the Global and Selective Attention Network (GSANet). 
It enhances all components of the SOTA DeepLabV3+ architecture using attention modules. The multiscale aggregation in the ASPP takes into account the importance of the contextual information using the proposed selective attention which deploys condensation and diffusion modules. More relevant global information is extracted using the proposed sparsemax global attention feature. The decoder also deploys attention with condensation and diffusion to dope  its different input features  with extrinsic information from the other  features.  With both the low-complexity MobileNetEdge FXN and the strong Xception FXN,  we show that GSANet  gives better performance than DeepLabV3+ and achieves the state of art accuracy. 
%
 %to improve the performance of semantic segmentation.
%In particular, two novel modules, namely the Selective Attention ASPP (SA-ASPP) and sparsemax based Global Attention Feature (sparsemax-GAF), are proposed.
%The SA-ASPP module is proposed to fuse multi-scale contextual information in ASPP, 
%While, the sparsemax-GAF module is applied to extract long range contextual information with sparse attentions 
%and replace the Global Average pooling in SA-ASPP.
%Extensive experiments show the effectiveness of our proposed method.
%In the future work, we will explore more lightweight design of these modules to further improve the computational efficiency while keeping the same level accuracy.

% \section{REFERENCES}
% \label{sec:ref}

\bibliographystyle{IEEEbib}
\bibliography{refs}

\end{document}